\documentclass[sigconf,screen=true,anonymous=false,bookmarks=false,nonacm]{acmart}
\usepackage{lipsum}
\usepackage{graphicx}
\usepackage{footnote}

\usepackage{amssymb}

\usepackage{footnote}
\usepackage{amsmath}
\usepackage{mathtools}
\usepackage{mathrsfs}                                      
\usepackage{siunitx} 
\usepackage{comment}
\usepackage{wrapfig} 
\usepackage[subrefformat=parens,labelformat=parens]{subfig}
\usepackage{bm,bbm}
\usepackage{multirow}
\usepackage{makecell}
\usepackage{threeparttable,booktabs}
\usepackage{blkarray}
\usepackage{tikz}
\usetikzlibrary{positioning,calc,fit,decorations.pathmorphing,shapes.geometric,shapes.gates.logic.US,calc}
\usepackage{tikz-3dplot}
\usepackage{balance}
\usepackage{courier}                                       
\usepackage{pifont}                                        
\usepackage{cleveref}                                      
\usepackage[mathcal]{eucal}
\usepackage[]{algpseudocode}                               
\algrenewcommand\textproc{\texttt}
\makeatletter\let\float@addtolists\relax\makeatother
\usepackage{algorithm}
\algblockdefx[pardo]{pardo}{endpardo}%
   [1]{\textbf{For} #1 \textbf{do in parallel}}%
   {\textbf{Endfor}}
\usepackage{filecontents}                                  
\usepackage{pgfplots}
\usepackage{pgfplotstable}
\usepackage{pgf-pie}
\usepgfplotslibrary{groupplots}
\pgfplotsset{compat=newest}
\usepackage[figuresright]{rotating}
\usepackage{xcolor,colortbl}                               
\usepackage[inline]{enumitem}                              
\setlist{leftmargin=5.08mm}

\newcommand{\minisection}[1]{\vspace{.06in}\noindent{\textbf{#1}}.}

\theoremstyle{plain}

\theoremstyle{definition}

\algrenewcommand\textproc{\texttt}

\usepackage[skip=1pt]{caption}            
\definecolor{CUHKorange}{RGB}{244,106,18} 
\definecolor{CUHKblue}{RGB}{0,111,190}    
\definecolor{CUHKgreen}{RGB}{0,127,128}   
\definecolor{CUHKred}{RGB}{228,46,36}     
\definecolor{CUHKyellow}{RGB}{198,148,34} 
\definecolor{CUHKdark}{RGB}{114,44,114}   
\definecolor{CUHKmiddle}{RGB}{144,44,144} 

\usepackage{tcolorbox}
\tcbuselibrary{skins,breakable}
    {\endtcolorbox}

\graphicspath{{./figs/}}

\begin{document}
\date{}

\title{
    Customized Retrieval Augmented Generation and Benchmarking for EDA Tool Documentation QA
}

\author{
    Yuan Pu$^{1,2}$, \quad
    Zhuolun He$^{1,2}$, \quad
    Tairu Qiu$^{2}$, \quad
    Haoyuan Wu$^{3}$, \quad
    Bei Yu$^1$ \\
    $^1$The Chinese University of Hong Kong, Hong Kong SAR \\
    $^2$ChatEDA Tech, China \\
    $^3$Shanghai AI Laboratory, Shanghai, China \\
}

\begin{abstract}
Retrieval augmented generation (RAG) enhances the accuracy and reliability of generative AI models by sourcing factual information from external databases, which is extensively employed in document-grounded question-answering (QA) tasks.
Off-the-shelf RAG flows are well pretrained on general-purpose documents, yet they encounter significant challenges when being applied to knowledge-intensive vertical domains, such as electronic design automation (EDA).
This paper addresses such issue by proposing a customized RAG framework along with three domain-specific techniques for EDA tool documentation QA, including a contrastive learning scheme for text embedding model fine-tuning, a reranker distilled from proprietary LLM, and a generative LLM fine-tuned with high-quality domain corpus.
Furthermore, we have developed and released a documentation QA evaluation benchmark, ORD-QA, for OpenROAD, an advanced RTL-to-GDSII design platform. 
Experimental results demonstrate that our proposed RAG flow and techniques have achieved superior performance on ORD-QA as well as on a commercial tool, compared with state-of-the-arts.
The ORD-QA benchmark and the training dataset for our customized RAG flow are open-source at \url{https://github.com/lesliepy99/RAG-EDA}.

\end{abstract}

\maketitle
\pagestyle{plain}

\section{Introduction}
\label{sec:intro}
Electronic design automation (EDA) comprises a set of software tools dedicated for the design, analysis and verification of electronic systems. 
To meet the increasing demand of semiconductor manufacturing and accommodate the advancing technology nodes, current EDA tools support a complex design flow with various functionalities and commands at all levels. 
Open-source EDA tools such as OpenROAD\cite{ajayi2019toward} and iEDA\cite{li2023ieda} provide complete RTL-to-GDSII flow and environment for circuit design. 
Commercial software, with rigorous consideration of industrial scenarios, supports much more sophisticated and various functionalities.
EDA tools are supposed to be equipped with well-organized and detailed documentations, yet
referring documentation often remains a pain for end users due to the huge amount of functionalities provided and the intertanglement among them.
To improve user experience and development efficiency of tool users, 
application engineers are employed and trained by vendors to provide on-site customer service and support.
Nevertheless, the high training and manpower cost has pushed people to consider an automated, no-human-in-the-loop means for such support. 

\begin{figure}
    \centering
    \includegraphics[width=.918\columnwidth]{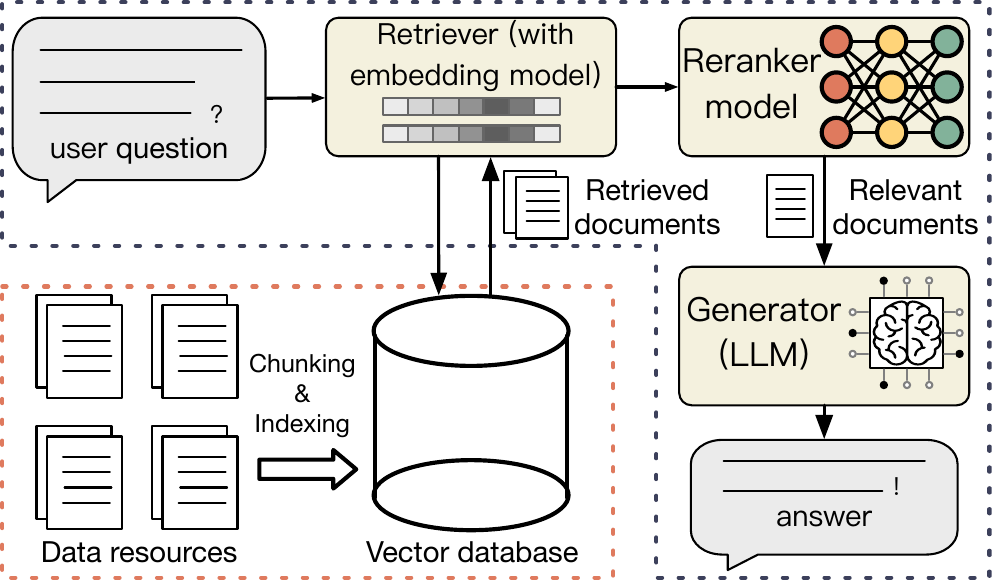}
    \caption{Illustration of the RAG flow.}
    \label{fig:rag}
\end{figure}

With the exceptional performance and rapid advancement of large language models (LLMs), retrieval augmented generation (RAG) has recently been proven to be an effective method for document grounded QA.
As is shown in \Cref{fig:rag}, to prepare for RAG, the entire document is segmented into reasonably-sized chunks, each of which is then embedded into a high-dimensional space by a text embedding model and stored in a vector database.
When a user poses a question, the text embedding model encodes the question into an embedding in the same space.
Through similarity matching, document chunks that most closely match the semantics of the question are retrieved as potential reference candidates.
Then, a reranker model conducts fine filtering and removes irrelevant document chunks from the candidates.
Finally, the relevant document chunks along with the user query are fed to the generator (usually an LLM) for answer generation.

During the process of RAG, the robust reasoning and inference capabilities of LLMs are effectively utilized, 
while the retrieved relevant information mitigates the hallucination issue and improves the reliability of LLM generation.
However, due to the mutually entangled structures of EDA tool documentations, sophisticated nature of EDA flow and rigorous logic reasoning involved in EDA-tool-related questions,
applying existing RAG flows for EDA tool documentation QA presents significant challenges.
The primary challenge is that existing RAG components lack expert knowledge in EDA, leading to inferior performance in EDA-specific contexts:
First, for retriever, existing text embedding models demonstrate poor semantic understanding of the terminologies and concepts of EDA, and are prone to retrieve irrelevant documents by similarity matching.
Second, existing reranker models fail to accurately differentiate between relevant and weakly-related documents under EDA-tool-specific scenarios.
For example, while the question is ``what is the command to place a specific pin'', the weakly-related document, which introduces the command to place all pins and is not helpful to answer the question, will be recognized as relevant by existing reranker models. 
 Third, existing generators (chat LLMs) lack EDA domain knowledge and demonstrate limited logical inference capabilities when addressing EDA-related questions, leading to inadequate answer generation.
Although many previous research work have customized LLMs for several EDA tasks including script generation~\cite{10299852,wu2024chateda}, HDL generation~\cite{liu2023chipnemo,chang2023chipgpt,liu2023verilogeval,fu2023gpt4aigchip, blocklove2023chip,thakur2023autochip,lu2024rtllm,thakur2023benchmarking,thakur2023verigen,pei2024betterv}, code verification and analysis~\cite{tsai2023rtlfixer,orenes2021autosva,kande2023llm,meng2023unlocking,paria2023divas,ahmad2023fixing,yao2024hdldebugger}, to the best of our knowledge, there has not been any research conducted on the documentation QA of EDA tools.

To resolve the challenges mentioned above, we propose \textbf{RAG-EDA}, a retrieval augmented generation (RAG) flow customized for EDA tool documentation question-answering. 
Our approach consists of three strategic enhancements: 
(1) For retriever, we first create a corpus of queries and answers that incorporate EDA domain knowledge and terminologies. 
Using this corpus, we employ contrastive learning to finetune a text embedding model, thereby enriching it with EDA-specific knowledge and enhancing its retrieval accuracy.
(2) For reranker, we distill the fine filtering capabilities of the proprietary LLM into our reranker model by contrastive learning, greatly improving its performance in weakly-related documents removal.
(3) For generator, we pretrain a chat LLM using the EDA domain corpus and subsequently finetune it using a dataset of EDA-tool-related QA pairs through instruction tuning. 

Furthermore, to assess the efficacy of our RAG flow and facilitate future research in EDA tool documentation QA, we have developed and made public a QA evaluation dataset, \textbf{ORD-QA}, based on the OpenROAD documentation.
This dataset comprises 90 high-quality question-document-answer triplets that span a variety of query types, including VLSI flow inquiries, command/option usage, installation guides, and GUI usage.
To further evaluate the robustness, universality and transferability of our proposed flow, we also design an internal QA evaluation dataset on the documentation of a commercial EDA tool.

Our major contributions are summarized as follows:
\begin{itemize}
    \item We carefully analyze the limitations of applying off-the-shelf RAG to EDA tool documentation QA, and propose a customized RAG flow to address them accordingly.
        Our solution outperforms for both academic and commercial EDA tools.
    \item We conduct contrastive learning and incorporate EDA expertise to finetune the text embedding model and the reranker model, achieving superior retrieval performance than SOTA retrievers.
    \item We propose a two-stage training scheme, namely, domain-knowledge pre-train and task-specific instruction tuning, to finetune an open-source chat LLM as the generator.
    \item We release a QA benchmark with $90$ high-quality QA pairs concerning OpenROAD, to evaluate our RAG flow and facilitate the future research on EDA tool documentation QA.
\end{itemize}

\section{Preliminaries}
\label{sec:prelim}

\subsection{Information Retrieval}
Given the user query, information retrieval (IR) is the process of obtaining information from a collection of documents or data sources that are relevant to the query.
Generally, there are two categories of methodology for information retrieval, namely, lexical (sparse) retrieval and semantic (dense) retrieval.

\minisection{Lexical (sparse) Retrieval}
In lexical (sparse) retrieval, the user's query is first tokenized into keywords. Techniques such as term frequency-inverse document frequency (TF-IDF) and BM25 are then used to assess the relevance of each keyword to each document chunk in the collection. The top-$k$ document chunks with the highest relevance scores are selected as the most relevant documents.

\minisection{Semantic (dense) Retrieval}
Given a user question and a list of document chunks, the approach of dense retrieval (semantic retrieval) utilizes the pretrained text embedding model to project the question and document chunks into the semantic vector space. The top $k$ document chunks which are the closest to the question in the vector space (thus have the most similar semantics with the question) are selected as the relevant documents \cite{karpukhin2020dense}.  

\minisection{Reciprocal Rank Fusion}
Reciprocal rank fusion (RRF)~\cite{cormack2009reciprocal} is a method used for data fusion in information retrieval systems. 
This technique combines the results from multiple retrieval systems (for example, lexical and semantic retrieval) to produce a single, enhanced ranking. 
Given a set of documents $D$ and a set of document rankings $R$ (each $r\in R$ ranks the relevance of $D$ for a specific retrieval system, and $r$ is essentially a permutation on $1\ldots |D|$), the RRF score of one document $d$ can be calculated as:
\begin{equation}
    \label{eq:rrf}
    \begin{aligned}
        \text{RRF}(d\in D)=\sum_{r\in R}\frac{1}{k+r(d)}, 
    \end{aligned}
\end{equation}
where $k$ is one constant and is usually set to 60.

\subsection{Performance Measurement}
\label{subsec:performance-metric}
To evaluate the effectiveness of a retriever/reranker in terms of its ability to retrieve all relevant documents/filter out irrelevant documents, we use recall as the metric.
Mathematically, recall@$k$ is defined as the proportion of relevant documents that are retrieved among the top $k$ results returned by the retriever/reranker.
Let $R$ be the set of all relevant documents in the corpus, and let $R_k$ be the set of relevant documents that appear in the top $k$  results returned by the retriever/reranker. 
Then, recall@$k$ is formulated as:
\begin{equation}
    \label{eq:recall}
    \begin{aligned}
        \text{recall}@k = \frac{|R_k|}{|R|}.
    \end{aligned}
\end{equation}

To evaluate the performance of a generator (a chat LLM) in answering EDA-tool questions,
we use three most popular metrics applied in the scenario of text generation and question-answering, namely, BLEU~\cite{papineni2002bleu}, ROUGE-L~\cite{lin2004rouge}, and UniEval~\cite{zhong-etal-2022-towards}.
These three metrics are used to measure the coherence, conciseness and factual consistency between the generated answer and ground truth answer for each question.

\begin{figure*}
    \centering
    \includegraphics[width=.918\linewidth]{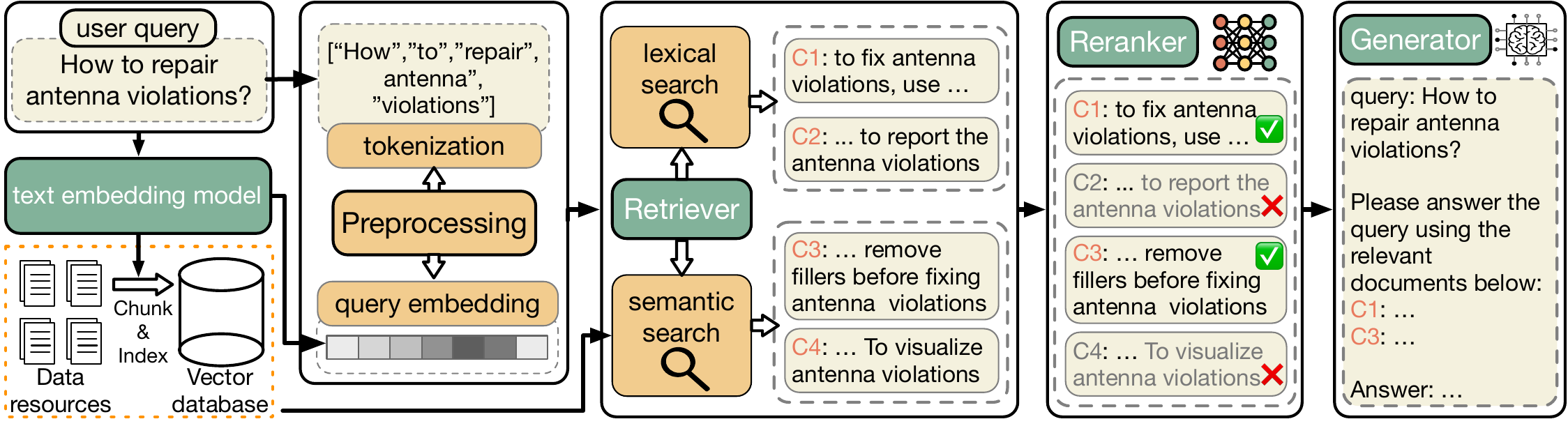}
    \caption{Overview of RAG-EDA, our proposed RAG flow for EDA tool documentation QA.}
    \label{fig:flow}
\end{figure*}

\minisection{BLEU}
Bilingual evaluation understudy (BLEU) is originally designed for measuring machine translation, but currently is widely used for the measurement of text generation. 
BLEU computes the precision of n-grams in the candidate answer by determining the proportion of n-grams that appear in the reference answer.
This is adjusted by a brevity penalty to discourage overly short translations. The metric is calculated as:
\begin{equation}
    \label{eq:bleu}
    \begin{aligned}
        \text{BLEU}=BP \cdot \exp \left( \sum_{n=1}^{N} w_n \log p_n\right),
    \end{aligned}
\end{equation}
which can be best understood as the geometric mean of multiple $n$-gram precision of various sizes.
Here, $BP$ (Brevity Penalty) is a multiplicative factor used to penalize short candidate answer, and $w_n$ are weights assigned to each logarithmic n-gram precision (we use uniform weights).
$N$ is the maximal n-gram order considered and $4$ is often a practical option.


\minisection{ROUGE-L}
ROUGE-L (recall-oriented understudy for gisting evaluation - longest common subsequence) measures the similarity between the answer generated by the LLM (denoted as $X$) and the reference answer (denoted as $Y$) by examining the longest common subsequence (LCS).
It computes both recall and precision based on the LCS, and it combines these into an F-measure. 
The formulas are defined as follows:
\begin{equation}
    \label{eq:rougeL}
    \begin{aligned}
        \text{recall}_{\text{LCS}} ={}& \frac{|\text{LCS}(X,Y)|}{|Y|}, \text{precision}_{\text{LCS}} = \frac{|\text{LCS}(X,Y)|}{|X|}, \\
        \text{ROUGE-L}             ={}& \frac{(1 + \beta^2) \cdot (\text{precision}_{\text{LCS}} \cdot \text{recall}_{\text{LCS}})}{(\beta^2 \cdot \text{precision}_{\text{LCS}}) + \text{recall}_{\text{LCS}}}, 
    \end{aligned}
\end{equation}
where $\beta$ is typically set to 1.

\minisection{UniEval}
UniEval~\cite{zhong-etal-2022-towards} is a widely-used LLM-based evaluation framework designed to assess the quality of generated text across multiple dimensions, including factual consistency, relevance, coherence, etc. 
Under the context of EDA tool documentation QA, we use the component of factual consistency score (ranging from 0 to 1, the higher the better) to measure the quality of the generated answer.

\section{Algorithms}
\label{sec:algorithm}

\Cref{fig:flow} illustrates the overall flow of \textbf{RAG-EDA}, our customized retrieval augmented generation (RAG) flow for EDA tool documentation QA.
In the preparation phase, the entire EDA tool documentation is segmented into reasonably sized chunks, which are then encoded into vectors by a customized text embedding model (introduced in \Cref{subsec:text-embedding-finetune}), forming a vector database.
The first stage is pre-processing, where the user query is tokenized into words and simultaneously encoded into an embedding.
The second stage, retrieval (introduced in \Cref{subsec:hybrid_retrieve}), operates in two parts: 
lexical retrieval (TF-IDF or BM25) and semantic retrieval using the pre-built vector database.
The results from both searches are combined to form relevant document candidates.
Subsequently, the finetuned reranker model (introduced in \Cref{subsec:rerank}) conducts fine filtering on the document candidates and further eliminates the weakly-related ones. 
Finally, the fine-filtered relevant document chunks along with the query are fed to the generator model (introduced in \Cref{subsec:generator}) for answer generation.
In the following sub-sections, we will detail the customization and technique for each stage.

\subsection{Domain-Customized Text Embedding}
\label{subsec:text-embedding-finetune}

\begin{figure}
    \centering
    \includegraphics[width=.938\columnwidth]{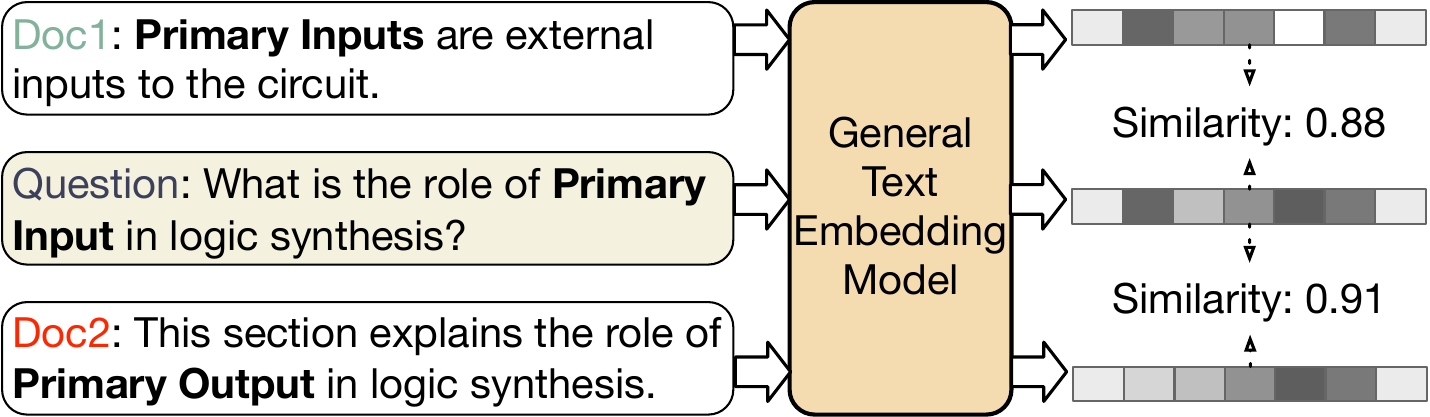}
    \caption{A fail case of general text embedding model in EDA-specific information retrieval (about \emph{Primary Input}). Doc1/Doc2 defines PI/PO. 
    The general text embedding model mistakenly perceives higher similarity between the question and Doc2.
    }
    \label{fig:embedding-retriever}
\end{figure}
During semantic retrieval, the text embedding model plays a crucial role in transforming both the user query and document chunks into numerical vectors that encapsulate their semantic content.
The accuracy of this semantic representation is essential as it directly impacts the retrieval performance. 
Although many proprietary and open-source text embedding models~\cite{xiao2023c,chen2024bge} perform well in general information retrieval tasks, they demonstrate poor semantic understanding of the terminologies and concepts in EDA.
As is illustrated in \Cref{fig:embedding-retriever}, the SOTA general-task text embedding model fails to adequately distinguish between basic EDA terms such as Primary Input (PI) and Primary Output (PO).
This deficiency leads to inaccuracies in the semantic vector space it constructs, potentially causing the model to erroneously select less relevant documents. For instance, despite Doc1 being more relevant to the query, Doc2 may be selected as the relevant document candidate due to its perceived semantic closeness, thereby compromising the overall effectiveness of the QA flow. 

\begin{figure}
    \centering
    \includegraphics[width=1\columnwidth]{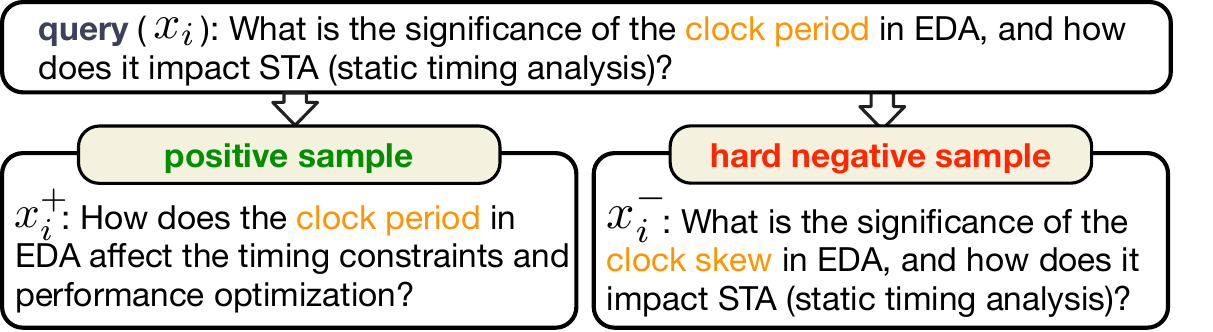}
    \caption{
    A contrastive data sample used for text embedding model finetuning.
    }
    \label{fig:pos_neg_sample}
\end{figure}

To address the aforementioned issue and enhance the performance of the retriever, we employ supervised contrastive learning to improve the model's ability to accurately perceive EDA terminologies and concepts.
The fundamental principle of our contrastive learning approach is as follows:
If two sentences involve the same EDA terminology, they should be positioned closely in the embedded vector space, despite differences in sentence structures.
Conversely, if two sentences have similar structures but pertain to different EDA terminologies or concepts, they should be distanced apart in the embedding space. 
Specifically, for one EDA-domain user query $x_i$, we denote its positive and hard negative samples by $x_i^+$ and $x_i^-$, respectively. 
The positive sample $x_i^+$ involves the same EDA term with the query $x_i$, meanwhile the negative sample $x_i^-$ has a similar sentence structure with the query, but pertains to different EDA terminology/concept. 
\Cref{fig:pos_neg_sample} shows one example of the triplet $\{x_i,x_i^+,x_i^-\}$: 
the query $x_i$ and the positive sample $x_i^+$ are user questions about "clock period", while the hard negative sample $x_i^-$ share the same sentence structure with $x_i$ but contains a different EDA terminology (clock skew).
We construct a triplet dataset $\{x_i,x_i^+,x_i^-\}_{i=1}^N$ of size $N$.
Then, we apply mini-batch gradient descent and supervised contrastive learning~\cite{gao2021simcse,zhang2023contrastive} to finetune the embedding model.
Assume the mini-batch size is $M$, 
and for each query $x_i$ in the mini-batch, we apply in-batch sampling. The contrastive loss of $s_i$ can be written as: 
\begin{equation}
    \label{eq:embedding-contrastive}
    \begin{aligned}
       -\log \frac{e^{sim(x_i,x_i^+)/\tau}}{\sum_{j=1}^M (e^{sim(x_i,x_j^+)/\tau}+e^{sim(x_i,x_j^-)/\tau})},
    \end{aligned}
\end{equation}
where $sim$ stands for the cosine similarity function, $\tau$ is the temperature hyper-parameter.

The text embedding model, finetuned through our supervised contrastive learning approach, exhibits improved semantic comprehension of the EDA-related corpus. 
This enhancement leads to superior document retrieval performance in the EDA-tool documentation QA task. 
Detailed results and analysis will be presented in \Cref{sec:experimental-results}.

\subsection{Hybrid Information Retrieval}
\label{subsec:hybrid_retrieve}
For EDA tool documentation QA, user queries span various categories including VLSI flow inquiries, command/option usage, installation guidelines, GUI usage and test unit, each characterized by distinct syntactic structure and sentence composition. 
Lexical (sparse) retrieval methods, such as TFIDF or BM25, are particularly effective when queries contain specific keywords or phrases directly matching the documentation content. For example, a query about the usage of a specific command like \emph{remove\_filler} is efficiently handled by these traditional lexical search techniques, which pinpoint the exact document section defining that command.
Conversely, when a query pertains to VLSI flow and does not necessarily contain direct keywords or phrases found in the documentation, semantic (dense) search method is employed. 
The semantic search focuses on matching the semantics between the query and the document chunks, thereby enhancing retrieval accuracy over lexical searches. 
For instance, a query like ``What are the steps for routing'' would enable a semantic search to retrieve document chunks that explain the commands \emph{global\_route} and \emph{detail\_route}.

Motivated by the preceding analysis, our approach deviates from the conventional methodology of solely employing semantic search for information retrieval~\cite{lewis2020retrieval}.
In our proposed framework, we implement a hybrid search strategy, integrating both semantic and lexical search techniques for the retrieval of documents.
Document chunks retrieved via the hybrid strategy are merged and deduplicated to form a pool of relevant document candidates. 
Subsequently, candidate documents that are weakly related to the query will be further differentiated and excluded, as detailed in \Cref{subsec:rerank}.

\subsection{
Reranker Finetuning}
\label{subsec:rerank}
\begin{figure}
    \centering
    \includegraphics[width=1\columnwidth]{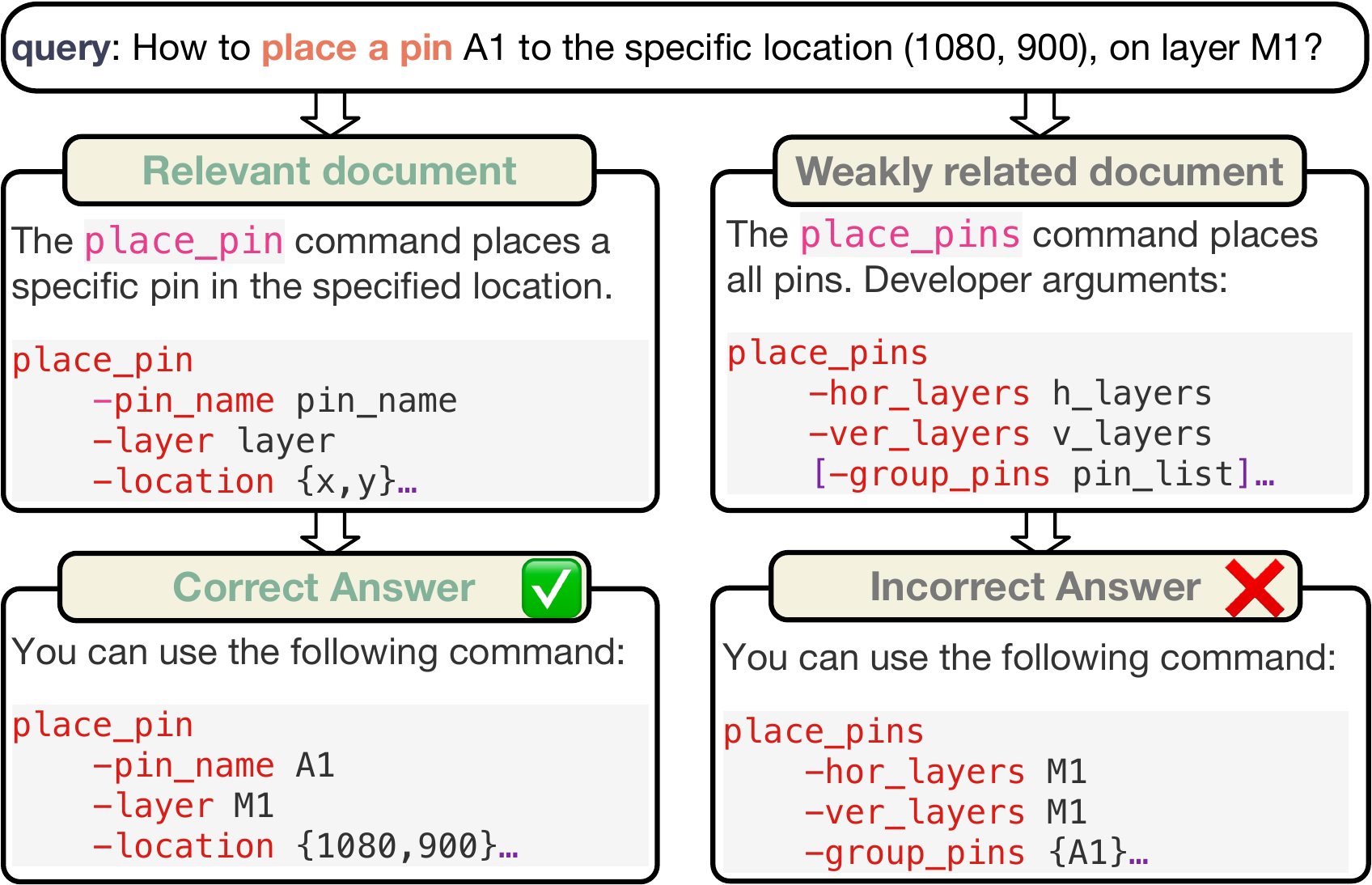}
    \caption{
        Weakly-related documents do harm to answer generation quality for EDA-tool questions.
    }
    \label{fig:weakly-related-doc}
\end{figure}
Candidate document chunks obtained by the hybrid retriever 
either share semantic similarities or contain the same keywords/phrases as the query.
Based on the relevance to the query, each candidate document chunk can be classified into one of two categories: (1) Relevant document whose content can be directly used to answer the query; 
(2) Weakly-related document which may contain the same EDA terminology/concept as the query, but can not be used to answer the query.
Previous research~\cite{cuconasu2024power} indicates that weakly-related documents may cause augmented LLMs to generate error-prone responses.
\Cref{fig:weakly-related-doc} demonstrates the impact of a weakly-related document chunk on answer generation for EDA tool queries.
The user inquires the command to place a specific pin. If the relevant document of the command \emph{place\_pin} is referred to, the generator gets correct answer. Meanwhile, referring to the weakly-related document, that is, the command to place all pins, the generator leads to wrong answer.

Re-ranking is one common technique to filter out weakly-related and irrelevant documents, and is implemented by many frameworks such as LlamaIndex, LangChain and HayStack.
SOTA rerankers in the RAG framework typically employ a cross-encoder architecture, which processes the query and each candidate document as a pair. It calculates a relevance score for each pair using a cross-attention mechanism. The top $k$ documents with the highest relevance scores are selected as the relevant documents.
Although SOTA cross-encoder reranker models such as bge-reranker\footnote{https://huggingface.co/BAAI/bge-reranker-large} demonstrate superior performance in general tasks of information retrieval, 
they are less effective in distinguishing between relevant and weakly-related documents in the context of EDA tool documentation QA.
As is shown in \Cref{fig:weakly-related-doc}, for a query requesting a command to place a specific pin, the reranker incorrectly identifies the command $place\_pins$ as relevant.
We observe that with well-designed prompt, the proprietary LLMs such as GPT-4 perform well in filtering out weakly-related documents for EDA-tool-involved QA . 
However, the autoregressive nature of GPT-4 makes it slow for document filtering, and proprietary LLMs are also costly and challenging to deploy.

To bridge the gap between existing cross-encoder reranker models and proprietary LLMs in the context of EDA tool QA documents filtering, 
we initiate our approach by collecting a set of EDA-tool-related questions, denoted as $Q=\{q_1,q_2,...,q_n\}$.
For each $q_i\in Q$, we apply the hybrid retriever described in \Cref{subsec:hybrid_retrieve} to retrieve $k$ candidate documents $C_i$.
We then employ GPT-4 with task-specific prompt to differentiate relevant documents ($C_i^+$) and weakly-related documents ($C_i^-$), ensuring  $|C_i^+|+|C_i^-|=k$.
Next, to enhance the ability of the reranker model in distinguishing between relevant documents and weakly-related documents, even when they share similar semantics or identical keywords with the query, 
we introduce a contrastive learning scheme to fine-tune the reranker model using the documents filtered by GPT-4.
Specifically, for a question $q_i\in Q$, we sample one positive document $s_i^+$ from $C_i^+$ and $m$ negative documents $\{s_{i,1}^-, s_{i,2}^-, ... ,s_{i,m}^- \}$ from $C_i^-$ ($m\leq k$), the contrastive loss for $q_i$ can be then written as:
\begin{equation}
    \label{eq:reranker-contrastive}
    \begin{aligned}
       -\log \frac{e^{f(q_i,s_i^+)/\tau}}{e^{f(q_i,s_i^+)/\tau}+\sum_{j=1}^m e^{f(q_i,s_{i,j}^-)/\tau}},
    \end{aligned}
\end{equation}
where $\tau$ denotes the temperature, $f(q,s)$ denotes the scalar score of the query-document pair $(q,s)$ outputted by the reranker model,
and the negative sample size $m$ is set to 3.

After being fine-tuned on the high-quality dataset, our reranker model exhibits high retrieval recall in EDA-tool QA scenarios. Detailed experiments and their results are discussed in \Cref{sec:experimental-results}.

\subsection{Domain-Specific LLM Generator}
\label{subsec:generator}
\begin{figure}
    \centering
    \includegraphics[width=1\columnwidth]{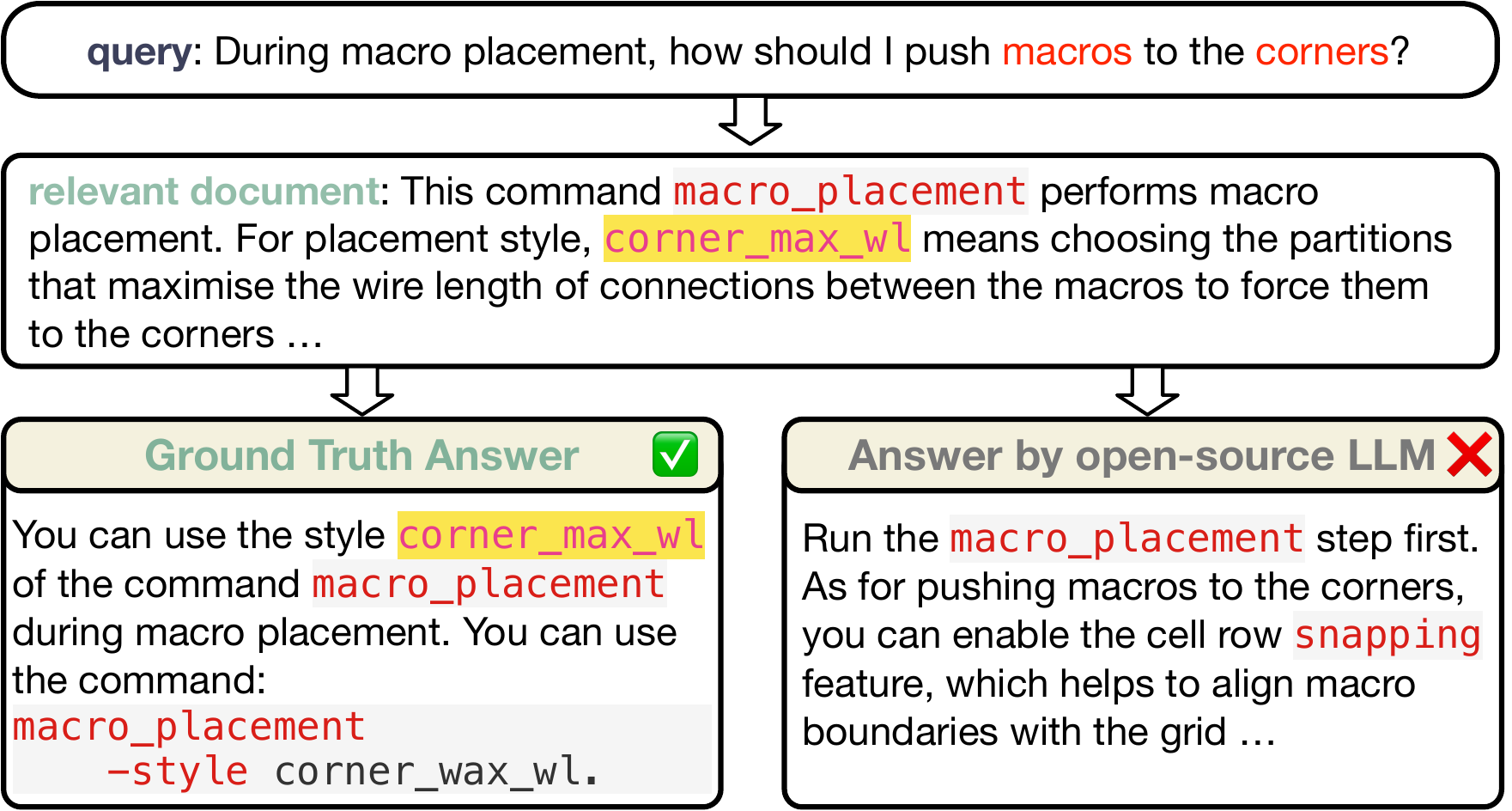}
    \caption{A fail case of existing open-source chat LLMs for answering the EDA-tool questions.}
    \label{fig:generator_eval}
\end{figure}

Serving as the last stage of the retrieval augmented generation (RAG) flow, the generator, usually an LLM, treats the user query and the filtered relevant document(s) as input, and output the answer.
We observe that directly applying the open-source chat LLMs as generators for EDA tool documentation QA leads to inferior quality of the answers. 
\Cref{fig:generator_eval} shows one typical example: The user inquires for the instruction to push macros to the corner during macro placement, the ground-truth answer to this question is to enable the \emph{corner\_max\_wl} style of the command \emph{macro\_placement}, which is depicted in the relevant document.
When using an open-source chat LLM, namely, Qwen-14B-chat, the generated answer is erroneous.

By analysis, we get the conclusion that there are mainly two limitations of existing chat LLMs for EDA tool documentation QA: 
1). Open-source chat LLMs lack expert knowledge in EDA. Questions involving the usage of EDA tools are usually complicated, besides the provided relevant document, some extra domain knowledge in EDA may be helpful in answering the question. For example, besides the description of the command \emph{macro\_placement}, the domain knowledge of VLSI macro placement is also helpful to answer the question in \Cref{fig:generator_eval}.
2). General chat LLMs are ineffective in handling the questions involved with EDA tools. 
Answering such questions requires intricate and domain-specific analysis between the question and provided documents, and step-by-step inference, which are not well mastered by existing chat LLMs.

\begin{figure}
    \centering
    \includegraphics[width=1\columnwidth]{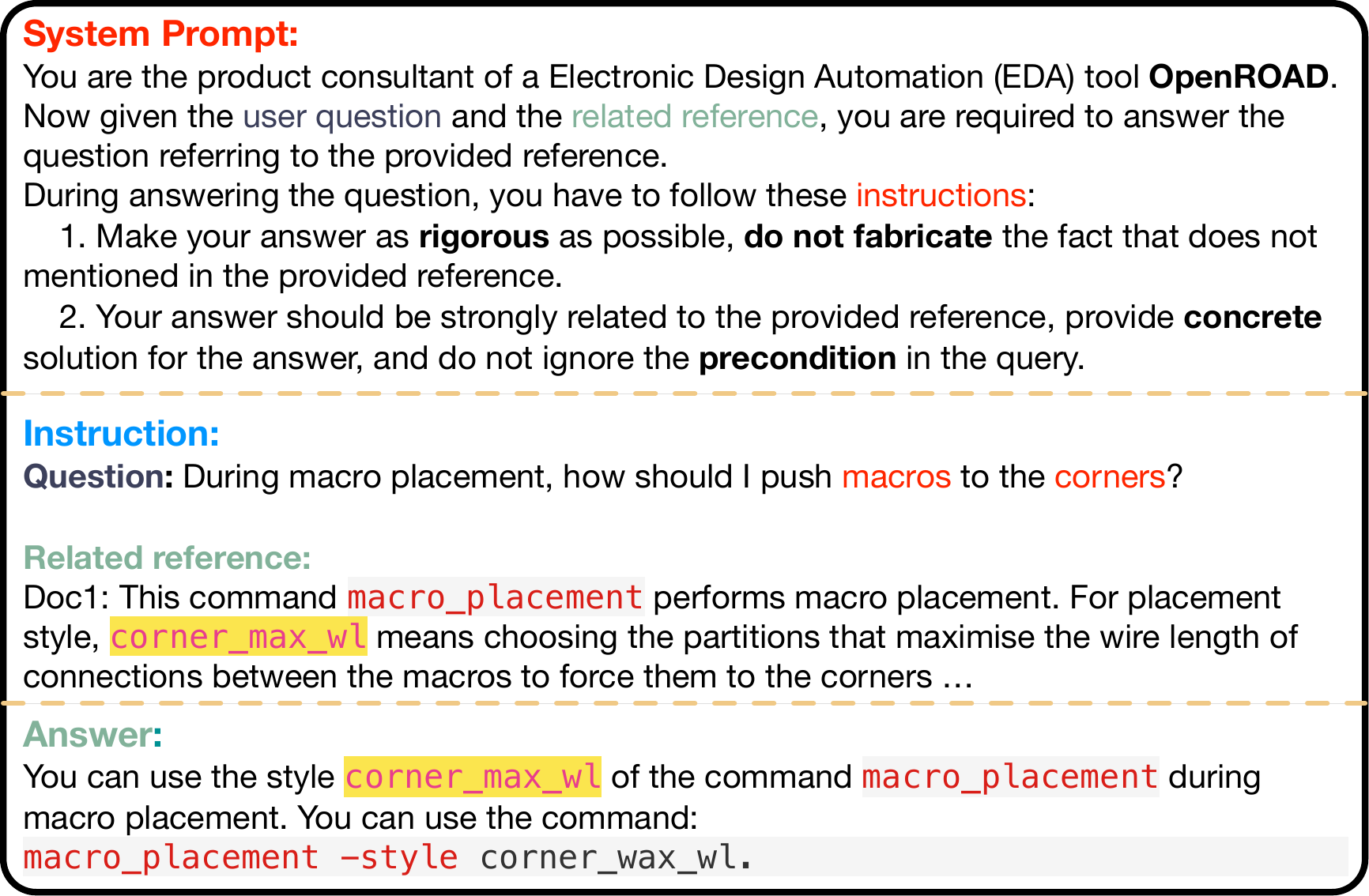}
    \caption{One sample of the QA data instance used from instruction tuning.}
    \label{fig:prompt_sample}
\end{figure}

To guarantee better quality of answer generation, we propose a two-stage training scheme to finetune the chat LLM as the generator.
The first stage is domain knowledge pre-train: 
We prepare two well-known textbooks in EDA, namely, \emph{Electronic Design Automation for IC Implementation, Circuit Design, and Process Technology}~\cite{lavagno2016electronic} and \emph{Handbook of Algorithms for Physical Automation}~\cite{alpert2008handbook}, as pre-train corpus. 
We divide the two textbooks into chunks by paragraph and construct a dataset, on which we pre-train the chat LLM in the autoregressive manner.
The second stage is task-specific instruction tuning:
To begin with, we leverage GPT4 to generate a set of triplets $\{q,r,a\}$ of high quality. 
For each triplet, $q$ is one question related to the usage of the EDA tool, $r$ is/are the document(s) used to answer the question, and $a$ is the groud-truth answer.
Then, we integrate a well-designed system prompt to each triplet to form a QA dataset. \Cref{fig:prompt_sample} shows one QA sample in the dataset.
Finally, we finetune the pre-trained chat LLM in stage 1 on the collected QA dataset.

Note that large language models (LLM) follow the autoregressive manner, which predicts tokens sequentially and the prediction of each token depends on the tokens that have been previously generated.
Given an LLM with parameter $\theta$ and a token sequence $x=\{x_1, x_2, ..., x_n\}$, the probability of the LLM to predict $x$ is formulated as:
\begin{equation}
    \label{eq:llm-probability}
    \begin{aligned}
      P(x) = \sum_{t=1}^{n} P(x_t | x_{<t};\theta).
    \end{aligned}
\end{equation}
The objective of LLM training is to minimized the negative log-likelihood loss formulated in \Cref{eq:llm-loss}: 

\begin{equation}
    \label{eq:llm-loss}
    \begin{aligned}
      L = - \sum_{t=1}^{n} \log P(x_t | x_{<t};\theta).
    \end{aligned}
\end{equation}
The two stages, namely, domain-knowledge pre-train and task-specific intrudction tuning, are trained using the loss function defined in \Cref{eq:llm-loss}, on their respective corpus dataset.

\section{Benchmark}
\label{sec:benchmark}

\begin{figure}
    \centering
    \includegraphics[width=0.868\linewidth]{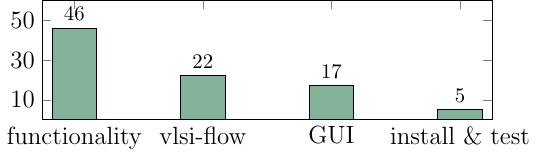}
    \caption{Distribution of question categories in ORD-QA.}
    \label{fig:question_category}
\end{figure}

To systematically and quantitatively evaluate the performance of RAG-EDA, 
we created and published the \textbf{ORD-QA} dataset,
which is open-source at \url{https://github.com/lesliepy99/RAG-EDA}.
ORD-QA contains 90 high-quality question-document-answer triplets based on the OpenROAD~\cite{ajayi2019toward} documentation.
For developing the dataset, 
we started by segmenting the complete OpenROAD documentation\footnote{https://openroad.readthedocs.io/en/latest/} into 290 chunks, based on the markdowns of sections and subsections.
Following the segmentation, each time we randomly sampled 1 to 5 document chunk(s).
If only one document chunk is sampled or there is logical relationship among the sampled chunks, 
we prompted GPT-4 to formulate a EDA-related question based on the document chunk(s), 
and analyzed the question alongside the provided document chunks to generate corresponding answer.
Each generated question-document-answer triplet underwent a rigorous manual review to correct any logical inconsistencies and fallacies.
Additionally, we manually crafted realistic user queries about the OpenROAD documentation and provided accurate answers to further enhance the dataset. 
The final ORD-QA dataset contains 90 high-quality <question, reference documents, answer> triplets.

The questions in the ORD-QA dataset are categorized into four types: functionality, VLSI-flow, GUI, and installation \& test. 
 Functionality questions detail the use of specific commands or options in particular user scenarios.
VLSI-flow questions request the steps to complete a given VLSI process. 
GUI questions address the usage of the OpenROAD GUI, while installation \& test questions cover software installation guidelines and unit test module inquiries.
\Cref{fig:question_category} shows the number distribution of different question categories in ORD-QA.
Most of the questions (82.2\%) are formulated based on 1 or 2 document chunks. The rest (17.8\%) are based on more than 3 document chunks. Among the 90 questions, 89 of them refer to no more than 5 document chunks, with only one question based on 7 document chunks.

\section{Experimental Results}
\label{sec:experimental-results}

\subsection{Training Dataset Collection}
\label{subsec:dataset}
This subsection discusses the process of collecting training dataset for the fine-tuning of all models used in RAG-EDA.

\minisection{Dataset for Text Embedding Model}
To construct the EDA-domain-specific <query, positive sample, negative sample> triplets mentioned in \Cref{subsec:text-embedding-finetune}, 
we first collected a comprehensive list of 261 EDA terminologies covering areas like logic synthesis, power analysis, partitioning, floorplanning, placement, routing, STA, DRC, and LVS.
Our dataset collection process involved selecting a terminology $t_i$ 
from this list and using GPT-3.5 to generate a related user query ($s_i$).	
GPT-3.5 also produced the answer for ($s_i$), which is used as the positive sample ($s_i^+$). 
Simultaneously, we created a negative sample ($s_i^-$) by replacing $t_i$ in $s_i$ with a different terminology $t_j$
 resulting in a data triplet $\{s_i,s_i^+,s_i^-\}$.
Overall, we gathered 3,975 <query, positive sample, negative sample> triplets to fine-tune the text embedding model.

\minisection{Dataset For Generator}
There are two training phases for generator: domain knowledge pre-train and task-specific instruction tuning.
For the construction of pre-train dataset, 
we split the two textbooks~\cite{lavagno2016electronic,alpert2008handbook} mentioned in \Cref{subsec:generator} into 4863 chunks,
where the token size of each chunk is limited within 1024.
The procedure of dataset collecting for task-specific instruction tuning is as follows: 
Each time we sample 10 document chunks of OpenROAD documentation as the document pool. Then, GPT-4 is prompted to select $n$ ($n \leq 10$) document chunks from the pool such that the selected chunks are logically related.
Based on the $n$ selected document chunks ($r$), we prompt GPT-4 to generate one question ($q$) from the perspective of the OpenROAD users, analyze the question step by step, and give the answer ($a$) referring to $r$.
Following the procedure above, we generate 1732 $\{q,r,a\}$ triplets as our dataset for instruction tuning.

\minisection{Dataset for Reranker Model}
To construct the dataset of reranker finetuning, we first randomly sample triplets from the instruction-tuning dataset.
For each question $q_i$, we treat their corresponding relevant document chunks $r_i$ as the positive documents and denote them by $C_i^+$.
Then, we apply hybrid retrieval described in \Cref{subsec:hybrid_retrieve} on $q_i$, and set the top-$k$ values for both the lexical and semantic searchers to be 10, in result, 20 document chunks are retrieved and denoted as $C_i^{\prime}$.
Among the chunks in $C_i^{\prime}$, we prompt GPT-4 to detect the ones which are weakly-related/irrelevant with $q_i$, and denote them by $C_i^-$.
Finally, we get 208 $\{q_i,C_i^+,C_i^-\}$ triplets as the dataset for reranker finetuning.

\subsection{Experimental Setting}
In our customized RAG flow, the text embedding model was fine-tuned on bge-large-en-v1.5\footnote{https://huggingface.co/BAAI/bge-large-en-v1.5} for 2 epochs, with a batch size of 16, a maximum sequence length of 128, and a temperature of 0.05. 
The reranker, based on the bge-reranker-large model\footnote{https://huggingface.co/BAAI/bge-reranker-large} with the XLM-RoBERTa architecture~\cite{conneau2019unsupervised}, underwent fine-tuning for 30 epochs, with batch size 4 and maximum sequence length of 512.
The generator, powered by the Qwen-14b-chat model~\cite{bai2023qwen}, was trained in two stages:
For domain knowledge pre-train, we pretrained the model on the textbook chunks dataset described in \cref{subsec:dataset} for 2 epochs with the learning rate of 1e-5 and batch size of 1.
For task-specific instruction tuning, we finetuned the model using QLoRA\cite{dettmers2024qlora} for 4 epochs with the learning rate of 2e-4 and batch size of 1. The maximal sequence length is 4096 and the lora rank is 32.
All the models above were fine-tuned on 16xA100 with 40G memory each.

\begin{table}[tb!]
    \centering
    \caption{Performance of semantic search for text embedding model on ORD-QA.
    }
    \label{tab:embedding}
    \resizebox{1.06\linewidth}{!}{
    \renewcommand{\arraystretch}{1.1}
    \begin{tabular}{|c|cccc|}
    \hline
    Model type & recall@5 & recall@10 & recall@15 & recall@20\\
    \hline \hline
    text-embedding-ada-002 & 0.447 & 0.534 & 0.609 & 0.634 \\
    bge-large-en-v1.5 & 0.503 & 0.596 & 0.634 & 0.660 \\
    Our embedding model & \textbf{0.547} & \textbf{0.658} & \textbf{0.702} & \textbf{0.733}\\
    \hline
    \end{tabular}
    }
\end{table}

\begin{table}[tb!]
    \centering
    \caption{Performance of reranker for weakly-related documents filtering on ORD-QA.
    }
    \label{tab:reranker}
    \resizebox{1.06\linewidth}{!}
    {
    \renewcommand{\arraystretch}{1.1}
    \begin{tabular}{|c|ccccc|}
    \hline
    Model type & recall@1 & recall@2 & recall@3 & recall@4 & recall@5\\
    \hline \hline
    RRF & 0.248 & 0.342 & 0.391 & 0.435 & 0.484 \\
    bge-reranker-large & \textbf{0.360} & 0.441 & 0.484 & 0.497 & 0.522 \\
    Our reranker & 0.335 & \textbf{0.534} & \textbf{0.609} & \textbf{0.665} & \textbf{0.671}\\
    \hline
    \end{tabular}
    }
\end{table}

\begin{table*}[tb!]
    \centering
    \caption{Performance of the LLMs as generators on ORD-QA.
    }
    \label{tab:generator}
    \resizebox{0.918\linewidth}{!}{
    \renewcommand{\arraystretch}{1.00}
    \begin{tabular}{|c|ccc|ccc|ccc|ccc|}
    \hline
    \multirow{2}{*}{Model Type}  & \multicolumn{3}{c|}{ORD-QA\#functionality} & \multicolumn{3}{c|}{ORD-QA\#vlsi-flow} & \multicolumn{3}{c|}{ORD-QA\#gui \& install \& test} & \multicolumn{3}{c|}{ORD-QA\#all} \\ 
    \cline{2-13}
    &BLEU & ROUGE-L& UniEval  &BLEU & ROUGE-L& UniEval&BLEU & ROUGE-L& UniEval&BLEU & ROUGE-L& UniEval   \\
    \hline \hline
   
    GPT-4~\cite{achiam2023gpt} & 0.116 & 0.257 & 0.724 & 0.149 & 0.268 & 0.759 & 0.184 & 0.328 & 0.798 & 0.141 & 0.277 & 0.751 \\
    Qwen1.5-14B-Chat~\cite{bai2023qwen} & 0.077 & 0.189 & 0.604 & 0.099 & 0.175 & 0.513 & 0.127 & 0.272 & 0.739 & 0.095 & 0.206 & 0.615 \\
    llama-2-13B-chat~\cite{touvron2023llama} & 0.099 & 0.224 & 0.694 & 0.110 & 0.214 & 0.654 & 0.125 & 0.250 & 0.678 & 0.108 & 0.228 & 0.680  \\
    Baichuan2-13B-Chat~\cite{yang2023baichuan} & 0.089 & 0.244 & 0.743 & 0.073 & 0.260 & 0.781 & 0.092 & 0.281 & 0.740 & 0.086 & 0.257 & 0.751 \\
    RAG-EDA-generator (ours) & \textbf{0.150} & \textbf{0.319} & \textbf{0.788} & \textbf{0.188} & \textbf{0.326} & \textbf{0.798} & \textbf{0.224} & \textbf{0.374} & \textbf{0.809} & \textbf{0.177} & \textbf{0.334} & \textbf{0.795}  \\
    \hline
    \end{tabular}
    }
\end{table*}

\begin{table*}[tb!]
    \centering
    \caption{Performance of the RAG flows on ORD-QA.
    }
    \label{tab:rag-ORD-QA}
    \resizebox{0.918\linewidth}{!}{
    \renewcommand{\arraystretch}{1.00}
    \begin{tabular}{|c|ccc|ccc|ccc|ccc|}
    \hline
    \multirow{2}{*}{RAG Flow}  & \multicolumn{3}{c|}{ORD-QA\#functionality} & \multicolumn{3}{c|}{ORD-QA\#vlsi-flow} & \multicolumn{3}{c|}{ORD-QA\#gui \& install \& test} & \multicolumn{3}{c|}{ORD-QA\#all} \\ 
    \cline{2-13}
    &BLEU & ROUGE-L& UniEval  &BLEU & ROUGE-L& UniEval&BLEU & ROUGE-L& UniEval&BLEU & ROUGE-L& UniEval   \\
    \hline \hline
    Vanilla-RAG~\cite{lewis2020retrieval} & 0.088 & 0.204 & 0.635 & 0.089 & 0.206 & 0.669 & 0.139 & 0.258 & 0.726 & 0.101 & 0.217 & 0.665 \\ 
    RAG-fusion~\cite{cormack2009reciprocal,bruch2023analysis} & 0.080 & 0.193 & 0.624 & 0.088 & 0.202 & 0.681 & 0.150 & 0.274 & 0.698 & 0.099 & 0.215 & 0.656 \\
    HyDE~\cite{gao2022precise} & 0.080 & 0.188 & 0.603 & 0.075 & 0.185 & 0.658 & 0.130 & 0.255 & 0.712 & 0.091 & 0.204 & 0.643 \\
    llmlingua~\cite{jiang-etal-2023-llmlingua,jiang-etal-2023-longllmlingua,wu2024llmlingua2} & 0.056 & 0.179 & 0.565 & 0.048 & 0.162 & 0.586 & 0.091 & 0.221 & 0.658 & 0.062 & 0.185 & 0.593 \\
    ITER-RETGEN~\cite{shao2023enhancing} & 0.098 & 0.208 & 0.652 & 0.089 & 0.202 & 0.632 & 0.136 & 0.251 & 0.700 & 0.105 & 0.217 & 0.659 \\
    RAG-EDA+GPT-4 & 0.101 & 0.230 & 0.681 & 0.116 & 0.234 & 0.690 & 0.178 & 0.299 & 0.742 & 0.123 & 0.248 & 0.698 \\
    RAG-EDA (ours) & \textbf{0.119} & \textbf{0.281} & \textbf{0.699} & \textbf{0.147} & \textbf{0.269} & \textbf{0.746} & \textbf{0.166} & \textbf{0.302} & \textbf{0.776} & \textbf{0.137} & \textbf{0.283} & \textbf{0.729} \\
   
    \hline
    \end{tabular}
    }
\end{table*}
\begin{table}[tb!]
    \centering
    \caption{Performance of the RAG flows on QA dataset of the commercial EDA tool.
    }
    \label{tab:rag-commcercial}
    \resizebox{0.768\linewidth}{!}{
    \renewcommand{\arraystretch}{1.00}
    \begin{tabular}{|c|ccc|}
    \hline
    RAG Flow & BLEU & ROUGE-L & UniEval\\
    \hline \hline
    Vanilla-RAG~\cite{lewis2020retrieval} & 0.072 & 0.182 & 0.569\\
    RAG-fusion~\cite{cormack2009reciprocal,bruch2023analysis} & 0.157 & 0.275 & 0.663\\
    HyDE~\cite{gao2022precise} & 0.131 & 0.258 & 0.643\\
    llmlingua~\cite{jiang-etal-2023-llmlingua,jiang-etal-2023-longllmlingua,wu2024llmlingua2} & 0.078 & 0.216 & 0.553\\
    ITER-RETGEN~\cite{shao2023enhancing} & 0.163 & 0.283 & 0.676\\
    RAG-EDA+GPT-4 & \textbf{0.179} & 0.290 & 0.624\\
    RAG-EDA (ours) & 0.176 & \textbf{0.332} & \textbf{0.730}\\
    \hline
    \end{tabular}
    }
\end{table}
\subsection{Evaluation: Text Embedding Model}
In this sub-section, we evaluate the performance of dense (semantic) search by our finetuned text embedding model, on the benchmark ORD-QA. 
We first use the under-test text embedding model to encode the 290 document chunks in \Cref{sec:benchmark} into the vector database.
Then, each question $q$ in ORD-QA is encoded by the text embedding model to obtain its vector representation $e^q$, and the similarity search library Faiss~\cite{douze2024faiss} is used to search for the top-$k$ closest document chunks with $e^q$ in the vector space. 
We use the metric recall@$k$ (introduced in \Cref{subsec:performance-metric}) to measure the capacity of the text embedding model in retrieving relevant documents.
We adopt the text-embedding-ada-002 model by OpenAI and bge-large-en-v1.5 as baselines, and set $k$ to be 5, 10, 15 and 20 for evaluating recall@$k$. 
Experimental results in \Cref{tab:embedding} show that fine-tuned by our well-designed constrastive learning scheme on the EDA corpus, 
our fine-tuned text embedding model dramatically outperforms its base model (bge-large-en-v1.5) and the SOTA commercial text embedding model by OpenAI, on the task of retrieving relevant documents for EDA-tool-involved questions. 

\subsection{Evaluation: Reranker Model}
To evaluate the performance of our fine-tuned reranker model on filtering out weakly-related and irrelevant documents, for each question $q$ in ORD-QA, 
hybrid search (described in \Cref{subsec:hybrid_retrieve}) is first conducted.
We use Faiss and BM25 for the lexical and semantic searches, respectively, and set the search limit of each retriever to be 20.
After hybrid document search, we obtain 40 candidate relevant document chunks for $q$.
Then, we de-duplicate the candidates.
Next, the reranker model calculates the similarity score between $q$ and each candidate document chunk, and the top $k$ document chunks with the highest scores are selected as the relevant document chunks.
Subsequently, we can calculate recall@$k$ to measure the fine-filtering ability of the reranker.
The SOTA open-source reranker model bge-reranker-large, based on which our reranker is fine-tuned, is selected as one baseline.
We also compare our reranker model with RRF (introduced in \Cref{sec:prelim}) since it is used in RAG-fusion, one commonly-used RAG flow. 
\Cref{tab:reranker} lists the experimental results on ORD-QA: by setting $k$ from 1 to 5, our fine-tuned reranker model outperforms the baselines in filtering out weakly-related documents and preserve relevant documents, for EDA-tool-related questions and documents.
 
\subsection{Evaluation: Generator}
To evaluate the performance of our pre-trained and fine-tuned generator on the task of answering EDA-tool-involved questions, for each question $q$ in ORD-QA, we combine $q$ and its golden relevant documents $r$ into a pre-designed prompt (\Cref{fig:prompt_sample} shows the template of the prompt). The prompt is then fed to the generator for answer generation. 
To quantitatively measure the performance of the generator, for each question, the BLEU, ROUGE-L and UniEval scores between the generated answer and the ground-truth answer are calculated. 
We select OpenAI GPT-4 and several SOTA open-source chat LLMs as our baselines, and experimental results in \Cref{tab:generator} show that being trained by our proposed two-stage EDA-specific training scheme described in \Cref{subsec:generator}, our generator is equipped with domain knowledge in EDA and possesses the ability of answering questions related to EDA tool documentation, and outperforms both commercial and academic chat LLMs on the ORD-QA benchmark.

\subsection{Evaluation: RAG Flow}
Finally, we evaluate the performance of our customized RAG flow. 
During the implementation of RAG-EDA, we use Faiss and BM25 as the semantic and lexical search engines for the retriever, and set the top-$k$ value of each searcher to 20. The top-$k$ of the reranker is set to 5.
For answer generation, our generator model is quantized by 4-bit for efficient inference. 
We select 5 cutting-edge academic RAG flows as our baselines, 
and they mainly differs in the way to retrieve documents.
\textbf{Vanilla-RAG}~\cite{lewis2020retrieval} solely applies semantic retrieval.
\textbf{RAG-fusion}~\cite{bruch2023analysis} uses both semantic and lexical retrievals and applies RRF~\cite{cormack2009reciprocal} for documents fusion and re-ranking.
For \textbf{HyDE}~\cite{gao2022precise}, hypothetical documents are appended to the query for accurate semantic retrieval.
\textbf{LLMlingua}~\cite{jiang-etal-2023-llmlingua,jiang-etal-2023-longllmlingua,wu2024llmlingua2} renders compression for the user query and retrieved documents for key information extraction and efficient inference.
Finally, \textbf{ITER-RETGEN}~\cite{shao2023enhancing} synergizes retrieval and generation in an iterative manner.

For the implementation of the baselines, we use Faiss and OpenAI text-embedding-ada-002 model for semantic retrieval, BM25 for lexical retrieval, and GPT-4 as the generator. 
To ensure fair comparison, the top-$k$ values of semantic retrieval for vanilla-RAG, HyDE, Llmlingua and ITER-RETGEN are set to 5. For RAG-fusion, the top-$k$ values of semantic and lexical retrievers are set to 20, and RRF returns the top-$5$ highest-score documents as relevant.
The number of retrieve-generation iterations of ITER-RETGEN is set to 2.
For ablation study, we also replace the generator of RAG-EDA by GPT-4 as one baseline.
\Cref{tab:rag-ORD-QA} shows the experimental results on ORD-QA:
The questions in ORD-QA can be divided into four categories as mentioned in \Cref{sec:benchmark}, and since the numbers of GUI questions and install \& test questions are small, we combine them together as a new category.
Consequently, columns 2 to 4 in \Cref{tab:rag-ORD-QA} list results for the three categories of questions, and the last column shows the average results on the whole benchmark.
Note that the metrics of BLEU, ROUGE-L and UniEval are calculated between the ground-truth answers (provided in ORD-QA) and the generated answers by the RAG flow.
It is observed that compared with the question categories of functionality and vlsi-flow, 
the GUI \& install \& test questions are easier to solve in nature, on which all the RAG flows achieve better performance.
Our proposed RAG flow (RAG-EDA) achieves the best performance on all categories of questions, and the baseline "RAG-EDA+GPT-4" (replacing the generator with GPT-4 in our proposed flow) ranks second, 
demonstrating effectiveness of our proposed RAG flow and fine-tuned generator in solving documentation QA for EDA tools.
The performances of RAG-fusion, HyDE and ITER-RETGEN are similar and worse than our customized flow, 
indicating the necessity of RAG flow customization for EDA tool documentation QA.
The result of Llmlingua is the worst, the reason is that the compression LLM of Llmlingua demonstrates inferior performance on handling EDA-tool related questions and documents. After the process of prompt compression, the key information in both the question and the documents are missing, leading to bad quality for answering.

\begin{figure}
    \centering
    \hspace{-.4in}
    \includegraphics[width=0.998\columnwidth]{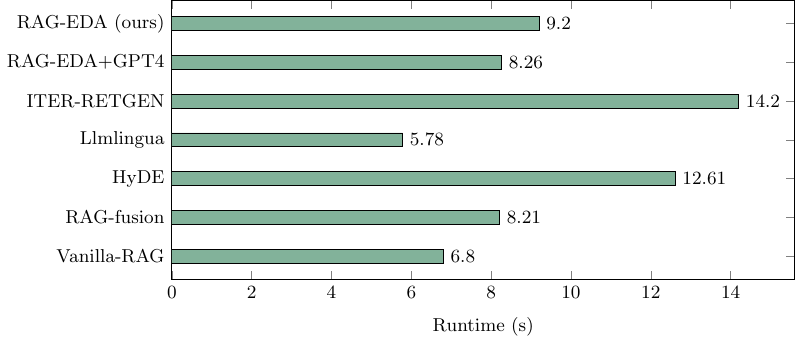}
    \caption{The average runtime  of the RAG flows to process one question in ORD-QA.}
    \label{fig:runtime}
\end{figure}

\Cref{fig:runtime} shows the runtime analysis of RAG-EDA and the baseline RAG flows on the ORD-QA benchmark. 
Llmlingua achieves the shortest average runtime, attributed to its use of compressed prompts which reduces the inference latency of the generator. Vanilla-RAG has the second shortest runtime due to its straightforward retrieval process.
The runtimes of RAG-EDA, RAG-EDA+GPT4 and RAG-fusion are similar, and the inference runtime of GPT-4 is slightly faster than our finetuned generator. 
HyDE and ITER-RETGEN take the longest runtime.
HyDE increases the generator's inference load by appending a hypothetical document to the query as a prompt, whereas ITER-RETGEN operates in an iterative manner, leading to higher computational costs.

To further verify the transferability and universality of our proposed RAG flow for different EDA tools, we collect a documentation QA evaluation dataset for a commercial EDA tool (the tool is a platform for timing ECO). The dataset follows the same standard as ORD-QA and contains 60 question-documents-answer triplets.
Experimental results in \Cref{tab:rag-commcercial} shows the outperformance of our proposed RAG flow compared with the baselines, thus certifies that our proposed RAG flow can be transferred to the task of documentation QA for different EDA tools.

\section{Conclusion}
\label{sec:conclusion}
We propose RAG-EDA, a customized retrieval augmented generation (RAG) flow for electronic design automation (EDA) tool documentation question-answering (QA). 
We develop domain-specific training strategies and use EDA-related corpus to customize the text embedding, reranker and generator models for RAG-EDA.
Besides, we also develop and release ORD-QA, a documentation QA evaluation benchmark based on OpenROAD.
RAG-EDA has achieved much better performance on both ORD-QA and a commercial EDA tool documentation QA benchmark compared with other state-of-the-arts.

\balance
\clearpage
{
\bibliographystyle{IEEEtran}

}


\end{document}